\documentclass[acmtog]{acmart}

\AtBeginDocument{%
  \providecommand\BibTeX{{%
    \normalfont B\kern-0.5em{\scshape i\kern-0.25em b}\kern-0.8em\TeX}}}

\title{Unrolled Primal-Dual Networks for Lensless Cameras}
\setcopyright{acmcopyright}
\copyrightyear{2022}
\acmYear{2022}
\acmDOI{10.1145/1122445.1122456}

\acmConference[Woodstock '18]{Woodstock '18: ACM Symposium on Neural
  Gaze Detection}{June 03--05, 2018}{Woodstock, NY}
\acmBooktitle{Woodstock '18: ACM Symposium on Neural Gaze Detection,
  June 03--05, 2018, Woodstock, NY}
\acmPrice{15.00}
\acmISBN{978-1-4503-XXXX-X/18/06}

\acmSubmissionID{papers\_325s2}

\DeclareMathOperator*{\argmin}{arg\,min}
\usepackage{makecell}

\begin{document}
\author{Oliver Kingshott}
\email{ucaboki@ucl.ac.uk}
\affiliation{%
  \institution{University College London}
  \city{London}
  \country{United Kingdom}
}

\author{Nick Antipa}
\email{naantipa@gmail.com}
\affiliation{%
  \institution{University of California San Diego}
  \city{San Diego}  
  \country{United States of America}
}

\author{Emrah Bostan}
\email{emrah.bostan@ams.com}
\affiliation{%
  \institution{ams OSRAM}
  \city{Lausanne}
  \country{Switzerland}
}

\author{Kaan Akşit}
\email{k.aksit@ucl.ac.uk}
\affiliation{%
  \institution{University College London}
  \city{London}
  \country{United Kingdom}
}

\renewcommand{\shortauthors}{Kingshott, et al.}

\begin{abstract}
Conventional image reconstruction models for lensless cameras often assume that each measurement results from convolving a given scene with a single experimentally measured point-spread function.
These image reconstruction models fall short in simulating lensless cameras truthfully as these models are not sophisticated enough to account for optical aberrations or scenes with depth variations.
Our work shows that learning a supervised primal-dual reconstruction method results in image quality matching state of the art in the literature without demanding a large network capacity.
This improvement stems from our primary finding that embedding learnable forward and adjoint models in a learned primal-dual optimization framework can even improve the quality of reconstructed images (+5dB PSNR) compared to works that do not correct for model error.
In addition, we built a proof-of-concept lensless camera prototype that uses a pseudo-random phase mask to demonstrate our point.
Finally, we share the extensive evaluation of our learned model based on an open dataset and a dataset from our proof-of-concept lensless camera prototype.
\end{abstract}

\ccsdesc[500]{Computing methodologies~Computational photography}

\keywords{lensless imaging, neural networks}

\begin{teaserfigure}
  \centering
  \includegraphics[width=\textwidth]{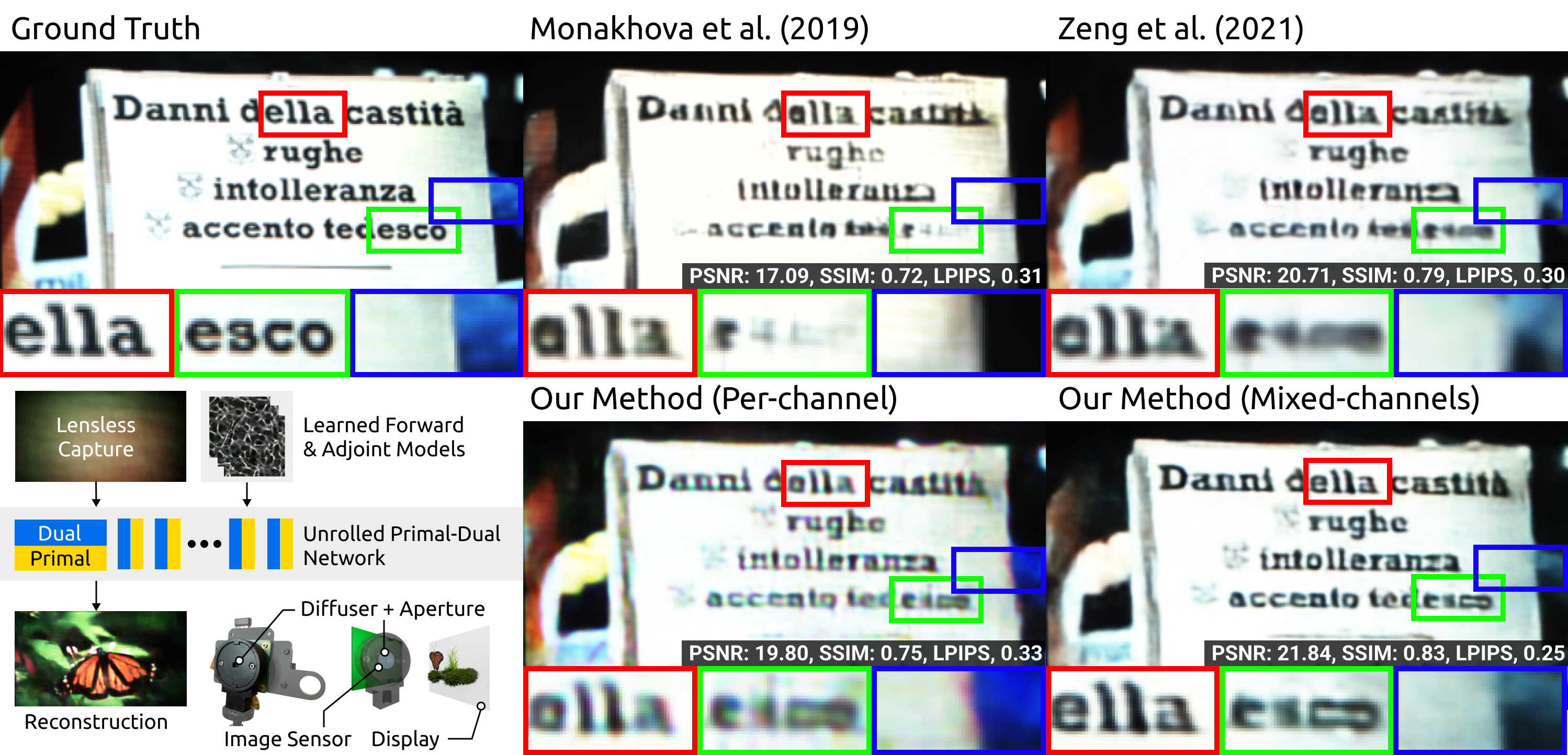}
  \caption{
   Comparison of our unrolled primal-dual network with state of the art.
   Intensive post-processing of lensless images cannot correct the model error, over-smoothing images and removing important features, such as text.
   We propose to replace classical lensless reconstruction methods with our physically-informed unrolled primal-dual model, where the model includes a series of learned forward and adjoint models (pseudo point-spread functions and their inverse).
   As a result, our work can produce plausible images and recover additional features while reducing the need for deep post-processing networks such as U-Nets~\cite{ronneberger2015u}
   (Source image courtesy MIR Flickr~\cite{huiskes2008mir}).
  }
  \label{fig:teaser}
\end{teaserfigure}

\maketitle
\section{Introduction}
A lensless camera uses a thin mask in place of a conventional lens.
Masks can manipulate phase, amplitude, or the entire complex light field of a given scene.
Unlike lenses in conventional cameras, these masks can be placed near the imaging sensor, enabling thinner and lighter imaging systems.
Additionally, lensless cameras offer the benefits of compressed imaging~\cite{fergus2006random,liutkus2014imaging}, embedding higher dimensional scene information such as depth from a single capture.
To benefit from these qualities, experts typically model lensless cameras as a linear system and recover images computationally by solving the inverse problem.

Pseudo-random phase masks have demonstrated adequate performance for lensless photography~\cite{Antipa2018,Boominathan2020}.
Unfortunately, image reconstruction typically requires computationally expensive and slow iterative reconstruction algorithms (e.g. ADMM~\cite{Antipa2018} and FISTA~\cite{Beck2009}).
To address this, a growing number of works use data-driven Convolutional Neural Networks (CNNs) to improve the speed and quality of lensless image reconstructions~\cite{Sinha17,Barbastathis19,Bae2020}.
A typical CNN with a limited receptive field size fails to accurately model the light transport of the imaging system~\cite{goodman2005introduction}, leading to learned models which fail to reconstruct lensless images accurately and efficiently.
Recent literature proposes neural networks that include a physical model with a large receptive field~\cite{Monakhova2019,Boominathan2020}.
These neural networks typically use a single-shot calibration measurement of the Point-Spread Function (PSF) to represent the physical model of the imaging system.
However, without the use of precisely engineered masks~\cite{Boominathan2020,tseng2021neural}, image formation in lensless cameras cannot be fully expressed by a single PSF model~\cite{yanny2020miniscope3d}.
This model mismatch can lead data-driven regularizers to hallucinate missing features or create overly smooth images.
Therefore, the development of models that can correct for model error without increased computational complexity or extensive calibration is of critical importance for the widespread adoption of lensless imaging.
Our proposed method replaces ADMM with a learned optimization scheme, improving image quality by reducing model error as opposed to intensive post-processing.
The result is a versatile deeply-calibrated lensless imaging architecture that avoids model error in the resulting reconstructions.
We provide the results of numerous experiments comparing our method against existing image reconstruction algorithms for lensless cameras.

Specifically, our work provides the following contributions:

\begin{itemize}
\item Learned primal-dual for lensless imaging.
We show for the first time that a modified learned primal-dual optimization framework~\cite{Adler2017} can recover images from a lensless camera using a pseudo-random phase mask.

\item Learned forward-adjoint model.
We embed additional linear operators within our learned primal-dual framework.
These learned forward-adjoint models are jointly optimized with the rest of our model using the same paired training examples. 
We show that our extended model provides a significant visual quality enhancement in our image reconstructions. 
Our method promises reductions up to 50\% in reconstruction error while using a fraction of the parameters compared to previous works.

\item Lensless camera prototype.
We build a proof of concept lensless camera to test further and demonstrate the performance of our model in an actual lensless camera with a pseudo-random mask. 
We provide an automatic calibration routine that can train our model without the need for an additional camera with a conventional lens.

\end{itemize}

\paragraph{Limitations}
When compared to models that use a single calibrated forward model, our method yields an improvement in the quality of lensless image reconstructions.
However, a thorough investigation is required to identify explainable links between our learned forward models and physically accurate models in the future.
In our experiments with our in-house built camera, we observe a lesser quality in image reconstructions when compared with the state of the art datasets~\cite{Monakhova2019,Boominathan2020}.
We believe these originate from the fact that the off-the-shelf diffuser we use does not fully resemble the case that we draw our inspiration from~\cite{Antipa2018}.
However, our work significantly improves the image quality both on benchmark datasets~\cite{Monakhova2019} and our in-house built camera.
\section{Related work}
We introduce a novel image reconstruction method for lensless cameras.
Here, we provide a brief survey of prior art in lensless cameras, unsupervised lensless image reconstruction methods and learned image reconstruction techniques.
Curious readers can read more about lensless cameras through the work by Boominathan et al.~\cite{boominathan2022recent}.

\subsection{Lensless cameras}
The idea of building cameras without requiring optical lenses has been a long-standing vision for scientists~\cite{barker1920pin} as optical lenses can be bulky, hard to manufacture with great precision, and are typically focused at one plane at a time.
The advent of ubiquitous high performance computing and the promise of high dimensional capture has led to a resurgence of interest in lensless cameras.
Mask based lensless cameras have been demonstrated with coded illumination~\cite{zheng2021coded}, coded apertures~\cite{horisaki2020deeply, Asif2017}, amplitude-only diffraction gratings (e.g., pinhole arrays~\cite{anand2020single}), photon sieves~\cite{yontem2018imaging}, separable amplitude masks~\cite{deweert2015lensless}, Fresnel Zone plates~\cite{wu2020single}), phase-only diffraction gratings~\cite{bernet2011lensless, Antipa2018} and metalenses~\cite{tseng2021neural}.
Additionally, the mask used in a lensless imaging system can also be co-designed with an algorithm that recovers scene information~\cite{tseng2021neural}.
The depth-varying PSFs of phase mask imaging systems can augment existing 2D imaging sensors with near-field 3D imaging~\cite{Antipa2018}.
Alternatively, single-pixel detectors combined with coded illumination patterns can be used for time-based imaging~\cite{Huang2013, Satat2017}.

In our work, we show a lensless camera prototype for experimental validation. 
Our prototype is similar to the one demonstrated by \citet{Antipa2018} but differs in implementation details, which we go through in our implementation section.

\subsection{Unsupervised Lensless Image Reconstruction Methods}
The large spatial extent of the PSFs used in phase-mask based lensless cameras necessitates a cropped convolution model, owing to the limited size of the imaging sensor.
By modelling the convolution and the sensor crop as separable sub-problems, the Alternating-Direction Method of Multipliers~\cite{Antipa2018} can be used to recover images using convex optimization.
However, modelling field-varying aberrations is cumbersome process using convex optimization approaches, typically requiring a 10x or greater increase in computational cost~\cite{yanny2020miniscope3d}.

\begin{figure*}
  \includegraphics[width=0.9\textwidth]{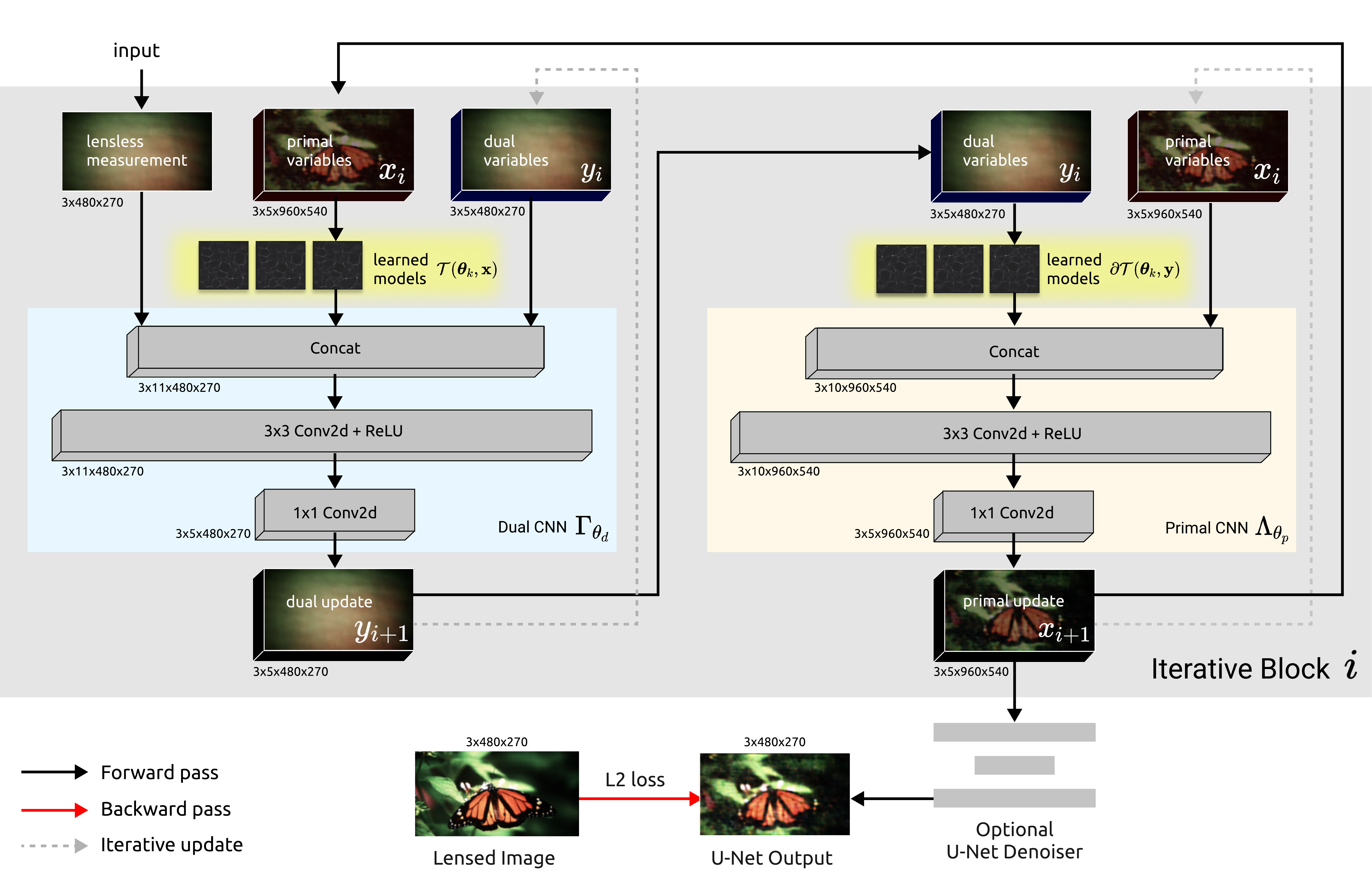}
  \caption{
  Unrolled Primal-Dual network architecture for reconstructing lensless images. 
  Our model accepts inputs in the form of a batch of RGB lensless measurements with a predetermined width and height. 
  The blue box illustrates our dual update step, where variables in the measurement domain ($\{y_i, y_{i-1}, b\} \in Y$) are concatenated channel-wise before passing through two convolutional layers parameterized by $\theta^i_d$. 
  The yellow box illustrates our primal update step, where each variable in the measurement domain ($\{x_i, x_{i-1}\}$) is likewise concatenated and convolved with two layers parameterized by $\theta^i_p$. 
  Our forward-adjoint model tensor, $\theta_k$, which is initialized with the value of a PSF measured using a point light source, is also optimized at each epoch. 
  Finally, our trained model reconstructs images from lensless measurements.
  }
  \label{fig:model}
\end{figure*}

\subsection{Learned Lensless Image Reconstruction Methods}
The advent of learning-based approaches eases the computational burden of lensless image reconstruction.
The work by~\citet{Monakhova2019} unrolls five iterations of ADMM and uses a large U-Net~\cite{ronneberger2015u} to improve perceptual quality.
However, this approach has a limited ability to correct for model error in the resulting reconstructions, relying on intensive post-processing to achieve plausible reconstructed images.
To our knowledge, the work by ~\citet{rego2021robust} demonstrates the first attempt at implementing a blind deconvolution model for lensless cameras without involving PSF measurements.
Our model requires re-training for each phase mask, yielding higher quality lensless reconstructions at the cost of portability.
\citet{Mitra2020} propose a fast learned reconstruction model for lensless cameras.
By improving boundary conditions inherent in the sensor crop, they show that they can recover realistic images in a single step without the need for an iterative model.
Our work embeds multiple large kernels within an unrolled iterative model to better compensate for optical aberrations.
\citet{Zeng2021} tackles model mismatch caused by imperfect modelling mainly due to spatially-varying PSFs with varying eccentricity.
They achieve this by learning residual blocks during each unrolled iteration of ADMM, which are fed into the U-Net denoiser to correct for model error.
We show that our method yields accurate intermediate reconstructions by separating the role of the denoising network from the model reconstruction network.
Most recently, \citet{Yanny2022} have proposed pairing multiple Wiener filters with convolutional neural networks to recover accurate images in a lensless microscopy application.
However, their method requires a experimental verification for phase-mask based lensless cameras as it targets microscopy.

In conclusion, existing learned methods depend both on accurate PSF calibration and additional training data to develop a suitable image prior.
Our method makes better use of supervision by diverting trainable parameters towards improving the underlying physical model of light transport.
By directly correcting for model-error, our method produces accurate intermediate reconstructions that are more consistent with images captured by a lensed camera.
To our knowledge, our learned method delivers results that are on-par with the current state of the art in terms of speed and image quality, while offering greater parameter efficiency than previous works.
\section{Method}
We first introduce the forward model for a phase-mask based imaging system.
We then present our proposed lensless image reconstruction model.
Finally, we illustrate our deep calibration procedure which captures the necessary dataset for our supervised model-based reconstruction.

\subsection{Problem Formulation}
We assume that measurements from our imaging system, $\mathbf{b}$, are the result of a linear transformation $\mathbf{A}$ applied to points in the scene $\mathbf{x}$, with some additional noise $\epsilon$:
\begin{equation}\label{imaging_system}
	\mathbf{b} = \mathbf{Ax} + \epsilon,
\end{equation}
where $\mathbf{b}$ and $\mathbf{x}$ are vectors.

Each column of $\mathbf{A}$ corresponds to the linear transformation of a single point in the scene, also known as PSFs.
Storing PSFs for each point in memory is a demanding task.
Rather than storing all PSFs, using an aperture enables the approximation of $\mathbf{A}$ as a cropped convolution with a PSF measured along the optical axis~\cite{Antipa2018}
\begin{equation}\label{psf}
	\mathbf{b} = \mathbf{C}(\text{PSF} * \mathbf{x}) + \epsilon
\end{equation}.
Here, $*$ represents a circular convolution and $\mathbf{C}$ represents a crop down to the size of the imaging sensor. 
The lateral shifting of the large PSF outside of the bounds of the image sensor necessitates this cropped convolution model.
A single experimentally measured PSF is typically used to reconstruct images using the described convolutional forward model~\cite{Antipa2018,Monakhova2019}.
The on-axis PSF is typically measured by shining a point light source along the optical axis of an existing system.
Under the assumption that $\mathbf{b}$ is the result of a cropped convolution with an experimentally measured PSF, we recover an estimate of the scene $\mathbf{x}$ by solving a regularized optimization problem:
\begin{equation}\label{variational}
	\mathbf{\hat{x}} \leftarrow
	\argmin_{\mathbf{x}}
    \frac{1}{2}\|\mathbf{C}(\text{PSF} * \mathbf{x}) - \mathbf{b}\|^2_2 + \lambda \mathcal{R}(\mathbf{x}),
\end{equation}
where $\mathcal{R}$ is a regularization function that penalizes unlikely solutions in the presence of noise, with $\lambda$ controlling the amount of regularization with respect to the data fidelity term.

In this work, we seek to improve the quality of lensless imaging by embedding learnable convolution kernels that are the same size as the PSF within a learned optimization scheme.

\subsection{Learning Large Kernels with Physically Informed Networks}
Access to paired training examples unlocks a vast landscape of learned reconstruction techniques.
Purely data-based architectures, such as U-Nets, typically require large numbers of paired training examples and suffer from poor generalization on unseen data.
These limitations can be overcome by incorporating knowledge of physical processes, such as light transport~\cite{kavakli2022learned}, into the neural network architecture.
Physically informed networks such as learned primal-dual~\cite{Adler2017} are highly data-efficient, requiring only a moderate number of training examples, and tend to generalize well to unseen data.
With access to paired training examples, but without knowledge of the true linear system $\mathbf{A}$, we propose to train a reconstruction network $\mathcal{G}$ with the goal of minimizing the average mean squared distance to ground truth reconstructions from a lensed camera $\mathbf{x_{gt}}$:
\begin{equation}\label{learned}
    \mathcal{L}_{MSE} := \|\mathcal{G_\theta}(\mathbf{b}) - \mathbf{x_{gt}}\|^2_2
\end{equation}
In the next section, we explain the design of $G_\theta$.
As the focus of our work is to recover the signal encoded in $\mathbf{b}$, we exclusively use mean-squared error as our loss function.

\subsubsection{Learned Primal Dual with a Physical Model}
We propose a modified learned primal-dual architecture as our learned reconstruction network $\mathcal{G}$  (Equation \ref{learned}).
Figure \ref{fig:model} illustrates how our data and parameters flow through the network.
We extend the original work by~\citet{Adler2017} in three ways.
First, we replace the forward operator $\mathcal{T}$ and its adjoint $\mathcal{\partial{T}}$ with the cropped convolution operation of our lensless camera in Equation (\ref{psf}):
\begin{equation}
\begin{split}
\mathcal{T}(x) &\leftarrow 
\mathbf{C}(\text{PSF} * x) \\
\mathcal{\partial{T}}(y) &\leftarrow \mathbf{P}(\text{PSF} \star y),
\end{split}
\end{equation}
where $\mathbf{P}$ represents zero padding up to twice the size of the imaging sensor, and $\star$ represents circular cross-correlation.
$x \in X$ and $y \in Y$ are primal and dual variables respectively, with the former belonging to the domain of reconstructed images $X$ and the latter in the domain of lensless measurements $Y$.

Second, we allow the PSF to be optimized during training.
We initialize $\theta_k \leftarrow \text{PSF}$, allowing the network to modify the physical PSF during training:
\begin{equation}
\begin{split}
\mathcal{T}(x) &\leftarrow 
\mathbf{C}(\theta_k * x) \\
\mathcal{\partial{T}}(y) &\leftarrow \mathbf{P}(\theta_k \star y),
\end{split}
\end{equation}

Finally, we wish to learn multiple kernels to improve our estimate of the true physical system.
We choose to learn $n$ convolution kernels, equal to the number of primal and dual variables. Let
\begin{equation}
\begin{split}
\mathbf{x} &= \begin{bmatrix}x^1 & x^2 & \dots & x^n\end{bmatrix} \\
\mathbf{y} &= \begin{bmatrix}y^1 & y^2 & \dots & y^n\end{bmatrix}, \\
\boldsymbol{\theta}_k &= \begin{bmatrix}\theta_k^1 & \theta_k^2 & \dots & \theta_k^n
\end{bmatrix},
\end{split}
\end{equation}
then each primal and dual variable $x^{1 \dots n}, y^{1 \dots n}$ is convolved or cross-correlated with its own learned kernel $\theta_k^{1 \dots n}$
\begin{equation}
\begin{split}
\mathcal{T}(\mathbf{x}) &\leftarrow \mathbf{C}(\boldsymbol{\theta}_k * \mathbf{x}) \\
\partial\mathcal{T}(\mathbf{y}) &\leftarrow \mathbf{P}(\boldsymbol{\theta}_k \star \mathbf{y}).
\end{split}
\end{equation}
The above modifications result in a variation of the learned primal-dual algorithm with the following update steps:
\begin{equation}\label{LPSFs}
\begin{split}
\mathbf{y}_i &\leftarrow \Gamma_{\theta^i_d}(\mathbf{y}_{i-1}, \mathcal{T}(\mathbf{x}_{i-1}), \mathbf{b}) \\
\mathbf{x}_i &\leftarrow \Lambda_{\theta^i_p}(\mathbf{x}_{i-1}, \partial\mathcal{T}(\mathbf{y}_{i})),
\end{split}
\end{equation}
where $\Gamma_{\theta^i_d}$, $\Lambda_{\theta^i_p}$ are small convolutional neural networks that are parameterized by each unrolled iteration $i \in {1 \dots 10}$.
At the end of the unrolled iterations, the variable $x_{10}^1$ is chosen as our best estimate of $\mathbf{\hat{x}}$.

\subsection{Per-channel \& Mixed-channel models}
To improve the performance of our method against baseline image quality metrics such as PSNR and SSIM, we propose an additional model based on higher dimensional feature maps as opposed to RGB images.
Specifically, we replace $k$ learned RGB kernels with $3 \times k$ single channel kernels, allowing for cross-channel communication across feature maps.
This results in a model with an increased signal-to-noise performance at the cost of a decrease in subjective color accuracy.
We provide a visual comparison of these two models and quantitative metrics in our results section.
\section{Implementation}
In this section we document the development of our own lensless camera as shown in Figure~\ref{fig:teaser}.
Additional details are provided in the supplementary material.

\paragraph{Camera Design}
We use a Raspberry Pi High-Quality camera connected to a Raspberry Pi Zero W.
This specific camera features a removable lens housing which we replaced with our own 3D printed design.
Following \citet{Monakhova2019}, we used a 0.5 degree engineered diffuser as our mask, placed ${\sim}10$mm away from the image sensor.
Our 3D printed housing is also illustrated in Figure \ref{fig:teaser}.
Our custom housing ensures that the optical element is placed at the desired distance from the imaging sensor, and contains space for an optional infrared filter.

\paragraph{Data Capture}
To capture a training and test dataset, we place our camera ${\sim}15$cm away from a 5.5 inch OLED display.
We illuminate a 5x5 square grid of pixels in the center of the display and capture the resulting image to measure the on-axis PSF.
We then use FISTA~\cite{Beck2009} to reconstruct a test image.
This test image is used to estimate a homography that warps each ground truth image to match the perspective of the lensless camera.
Automated software shows a variety of images from the DIV2K dataset~\cite{Timofte_2018_CVPR_Workshops}, capturing 8000 training images and 1000 test images.
\section{Evaluation}
We first present the results of comparing our method against two central state-of-the-art work that uses DiffuserCam dataset~\cite{Monakhova2019, Zeng2021}.
We additionally perform ablation studies to determine the contribution from each component in our method on reconstructed image quality.
Finally, we verify our method using our hardware prototype.

\subsection{DiffuserCam results}
We compare our model's results against the work that uses DiffuserCam dataset~\cite{Monakhova2019} in Table~\ref{tab:results}, where the number of parameters used, the size of training and testing examples, processing time, and image quality are considered.

\begin{table*}[h]
\begin{tabular}{llcrrrrrr}
 \toprule
 Method & PSFs & U-Net & PSNR & LPIPS & Parameters & Runtime (ms) & Training Examples & Iterations \\
 \midrule
 ADMM & 1 RGB (fixed) & & 11.97 & 0.60 & - & 1190 & 0 & 100 \\
 Le-ADMM & 1 RGB (fixed) & & 11.89 & 0.57 & 20 & 50 & 100 & 5 \\
 Le-ADMM-U & 1 RGB (fixed) & \checkmark & 20.46 & 0.37 & 10.6M &  55 & 24,000 & 5 \\
 \midrule
 Ours (RGB) & 1 RGB (learned) & & 16.74 & 0.54 & 0.4M & 74 & 9,000 & 10 \\
 & & \checkmark & 21.47 & 0.43 & 1.2M & 77 & 9,000 & 10 \\
 \midrule
 Ours (RGB) & 5 RGB (learned) & & 16.91 & 0.51 & 2.0M & 80 & 9,000 & 10 \\
 & & \checkmark & 23.48 & 0.40 & 2.7M & 88 & 9,000 & 10 \\
 \midrule
 Ours (Mixed) & 15 (learned) & & 19.00 & 0.48 & 2.0M & 82 & 9,000 & 10 \\
 & & \checkmark & 25.34 & 0.35 & 3.8M & 84 & 9,000 & 10 \\
 \bottomrule
\end{tabular}
\caption{Comparison of our models against previous work by \citet{Monakhova2019}.
Our model achieves produces modestly accurate reconstructions quickly without the use of a large U-Net, at the cost of learning additional large kernels $\theta^k$.
These kernels occupy the majority of our parameter space.
Adding a small U-Net to our models improves reconstruction quality further.
Increasing the number of learned kernels improves PSNR by ${\sim}2$dB when combined with U-Net denoising, with cross-channel denoising adding another ${\sim}2$dB.
}
\label{tab:results}
\end{table*}

\begin{figure*}
\includegraphics[width=\textwidth]{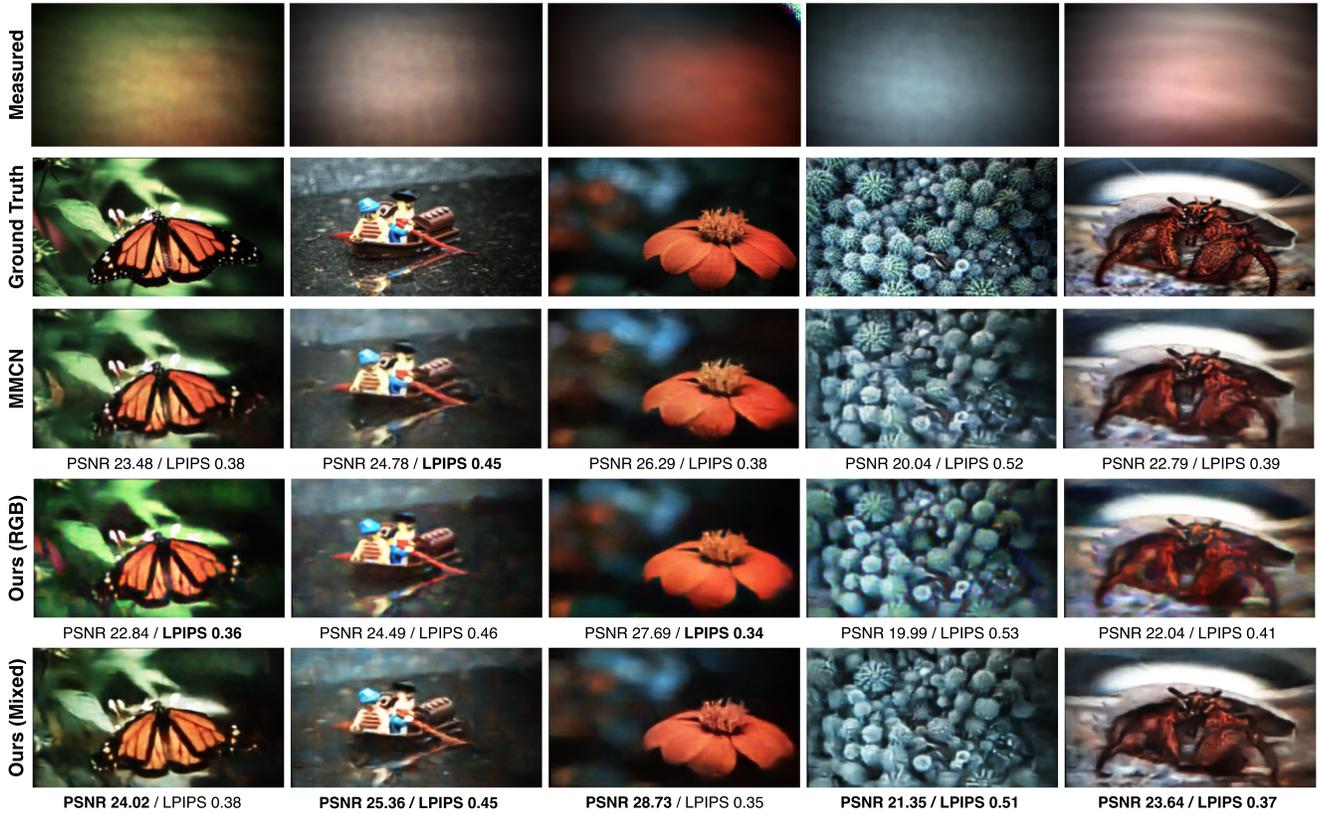}
\caption{Comparison of reconstructed test images against the ground truth images.
We compare our method against MMCN~\cite{Zeng2021}.
MMCN is based on five unrolled iterations of ADMM with additional residual blocks to correct for model error.
Our per-channel model (RGB) improves subjective color accuracy while our mixed-channel model (Mixed) recovers higher frequency content.
The primary feature of both models is that multiple large kernels are learned to correct for model error.
}
\label{fig:image-results}
\end{figure*}

Our results suggest that our proposed method improves the quality of images reconstructed from measurements captured by a lensless camera.
This is supported by qualitative results in Figure \ref{fig:image-results}, which appear to reproduce features that are more faithful to the original ground truth images.

\subsection{Ablation Studies}

\begin{figure*}
\includegraphics[width=\textwidth]{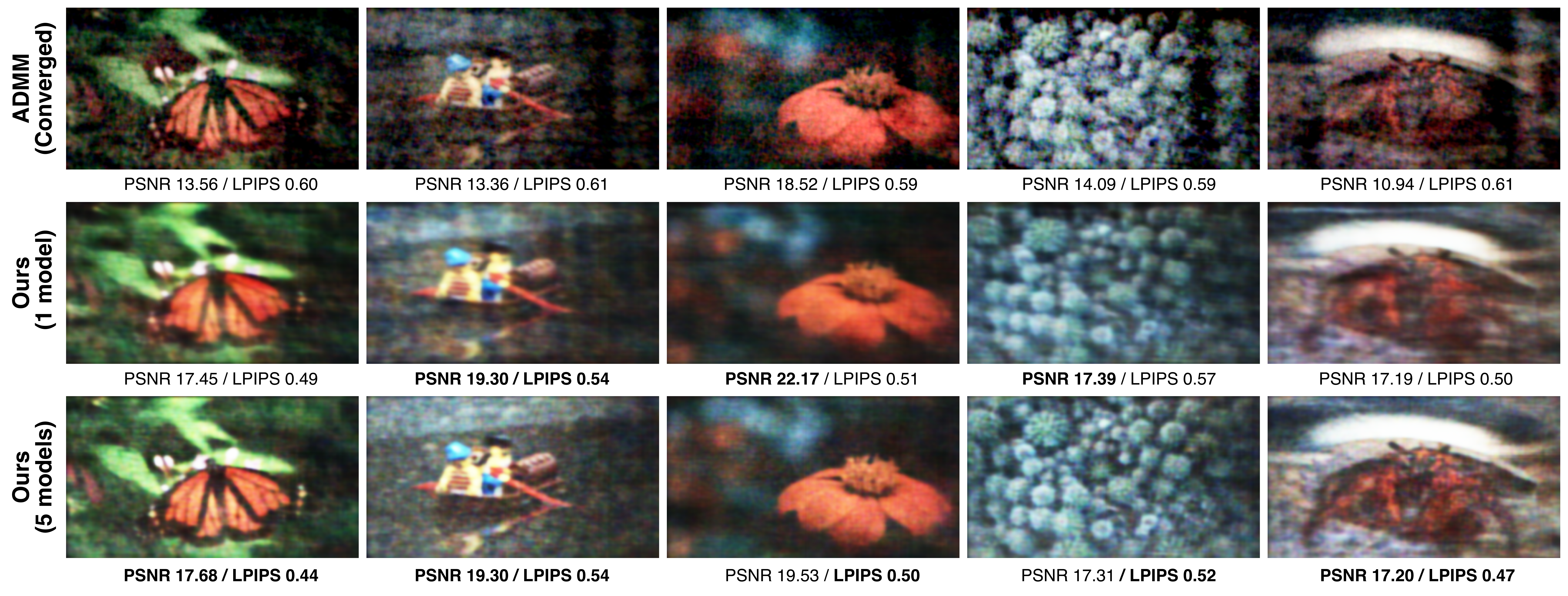}
\caption{Comparison of our learned model-based reconstruction networks against unsupervised ADMM (converged).
U-Net denoising was disabled to show that our intermediate reconstructions consist of images that are more faithful to the ground truth data.
Learning additional kernels appears to improve accuracy while yielding results faster than classical methods.
We reason that our network prioritises consistency with the true physical model, resulting in fewer artifacts.}
\label{fig:noisy-results}
\end{figure*}

\paragraph{Disabling U-Net Denoiser.}
To further confirm that the quality of our reconstructions has increased as a result of correcting for model error, we measure the quality of intermediate reconstructions without the use of a U-Net for denoising.
We show our qualitative results in Figure~\ref{fig:noisy-results} and quantitative results in Table~\ref{tab:results}.
When our U-Net is disabled, the resulting images are noisy but are faithful to the ground truth images.
Our intermediate reconstructions demonstrate that our model-based reconstruction network performs the bulk of the work in producing usable lensless reconstructions.

\paragraph{Effect of learning multiple models}
We ran an additional study to quantify the effect of decreasing the number of learned models from 5 to 1.
We include quantitative results in Table \ref{tab:results} and present reconstructed images from our reduced model in Figure \ref{fig:noisy-results}.
Decreasing the number of learned models from 5 to 1 decreases the resulting image quality after post-processing by ${\sim}2$dB.

\subsection{Prototype results}
\begin{figure}
\includegraphics[width=0.9\linewidth]{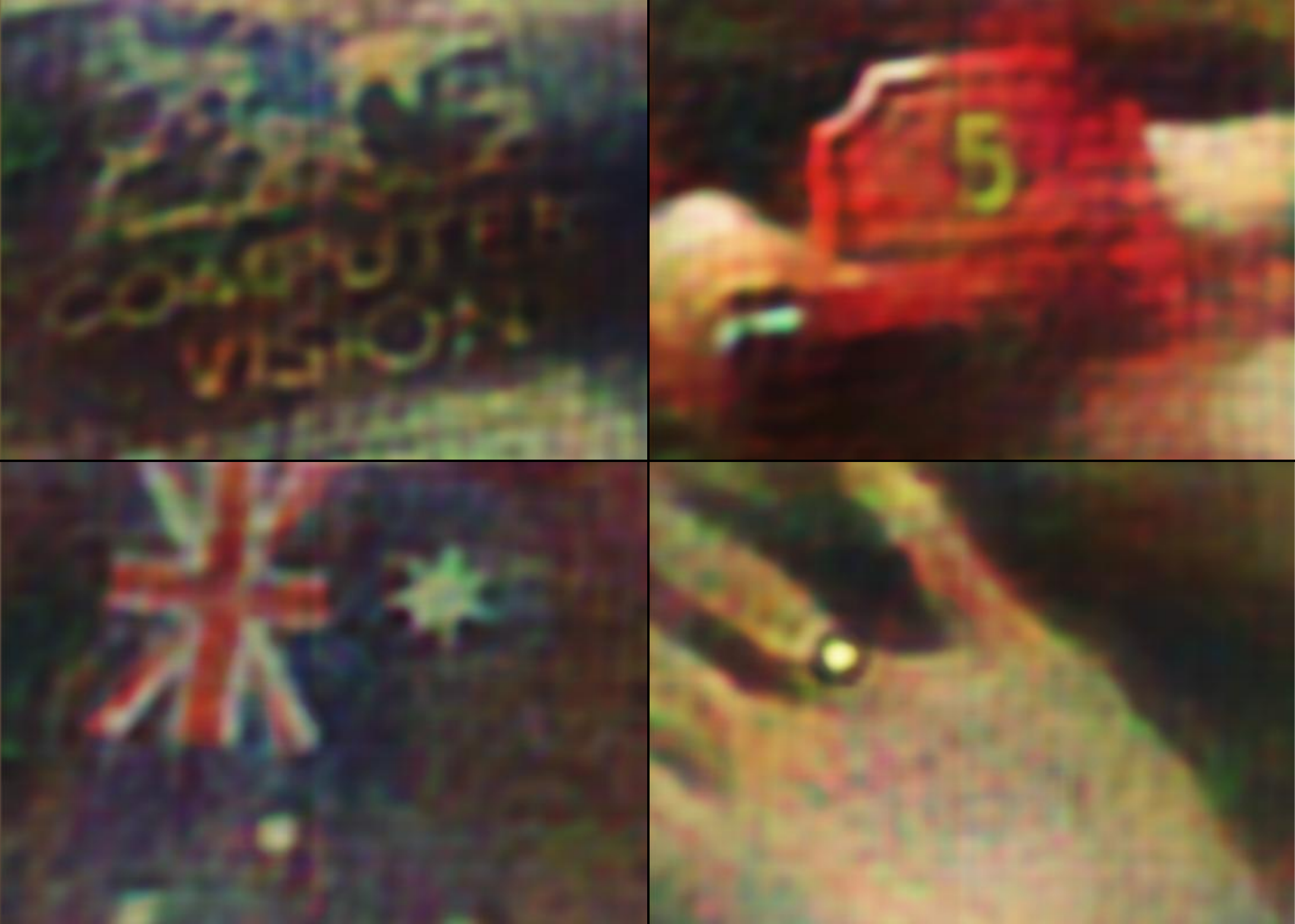}
\caption{Reconstructions from our lensless camera prototype trained using our RGB model.
Our optical element consists of a thick holographic diffuser (0.76mm) with bulk scattering, leading to a degradation in image quality.
}
\label{fig:wild}
\end{figure}

We additionally compare the results of our learned model using a prototype camera built in the lab.
We present sample reconstructions in Figure \ref{fig:wild} and provide additional reconstructions in our supplementary material.
\section{Discussion}
\label{sec:discussion}
\paragraph{Comparison to classical methods.}
Our proposed models are end-to-end differentiable.
They are trained to learn an unrolled iterative reconstruction algorithm, a physically informed model, and a suitable image prior.
While our model appears to produce accurate intermediate reconstructions, it is difficult to discretely map each learned component of the model to a specific component existing classical methods.
One line of future work could be to establish whether embedding learnable physical models within a classical variational method can achieve similar results.
A forward model that is learned independently of image priors and a chosen reconstruction algorithm could be used to evaluate the data fidelity of reconstructed images against their lensless measurements.

\paragraph{Comparison to learned methods.}
When compared to learned methods that use a fixed PSF calibration measurement, our method is able to reconstruct images that more closely resemble images captured by a lensed camera.
It is clear that the improved performance of our method is achieved by redistributing model parameters away from deep neural networks and towards the underlying physical model of light transport in lensless cameras.
However, the exact mechanism through which our model improves performance against existing learned methods is unclear.
It is possible that our model could be correcting for field-varying aberrations that are not captured by a single on-axis calibration measurement.
However, we note that our proposed methods lack any explicit mechanism to apply each learned model to a specific spatial region.
Finally, we note that our claim of improved data fidelity can only be measured implicitly by comparing our reconstructions with a lensed camera.
In future work, we would like to use measured or simulated field-varying PSFs to design robust models that can explicitly correct for field-varying aberrations without the need for manual calibration.

\paragraph{Color Accuracy.}
Our two proposed models highlight a potential trade-off between the recovery of high frequency details and color accuracy in phase mask cameras.
Allowing the mixing of color channels appears to increase the frequency content of recovered images.
However, our informal subjective opinion is that our per-channel model is able to reproduce color more accurately.
We suspect that our per-channel model is vulnerable to color fringing artifacts introduced by the chosen phase masks.
Future work could investigate treatment through the use of additional loss functions (such as those proposed by \citet{heide2013high}) or through improved phase mask design~\cite{Boominathan2020}.
\section{Conclusion}
Unconventional camera designs with thin masks in place of conventional lenses offer freedom from the constraints of traditional optics.
However, the speed of reconstruction and image quality in mask-based lensless camera designs remains a significant drawback.
We argue that neural networks with embedded physical priors for lensless imaging can help to counter this drawback.
We show that such an approach can provide on-par image reconstruction quality without demanding extensive resources in training.
Thus, we hope that our work can further develop performant and interpretable methods for lensless image reconstruction.

\section*{Acknowledgement}
We thank 
Laura Waller, Kristina Monakhova, Tianjiao Zeng and Edmund Lam for their support in providing useful insights from their work;
Tobias Ritschel for fruitful discussions at the early phases of the project; 
Koray Kavaklı for his support in hardware prototype related figure and camera homography related software; 
Tim Weyrich for dedicating GPU resource.
Kaan Akşit and Oliver Kingshott relied on the Royal Society's RGS\textbackslash R2\textbackslash 212229 - Research Grants 2021 Round 2 for building the hardware prototype.

\bibliographystyle{ACM-Reference-Format}
\bibliography{bibliography}
\end{document}